\def\year{2020}\relax
\documentclass[letterpaper]{article} 
\usepackage{aaai20}  
\usepackage{times}  
\usepackage{helvet} 
\usepackage{courier}  
\usepackage[hyphens]{url}  
\usepackage{graphicx} 
\usepackage{graphicx}  
\usepackage{booktabs}
\usepackage{amsmath}
\usepackage[]{algorithm2e}
\usepackage[skins]{tcolorbox}
\usepackage[]{todonotes}   
\usepackage{aurical}
\usepackage{calligra}
\usepackage{pbsi}
\usepackage[T1]{fontenc}

\newcommand{\citet}[1]{\citeauthor{#1} \shortcite{#1}}
\newcommand{\citep}{\cite}
\newcommand{\citealp}[1]{\citeauthor{#1} \citeyear{#1}}

\title{Learning to Learn Morphological Inflection for Resource-Poor Languages}
\author{Katharina Kann, Samuel R. Bowman \and Kyunghyun Cho\\
New York University, USA\\
\{kann, kyunghyun.cho, bowman\}@nyu.edu
}
 
\begin{document}

\maketitle

\begin{abstract}
We propose to cast 
the task of morphological inflection---mapping a lemma to an indicated inflected form---for 
resource-poor languages as a \textit{meta-learning} problem.
Treating each language as a separate task, we use data from high-resource source languages to learn a set of model parameters that can serve as a strong initialization point for fine-tuning on a resource-poor target language.
Experiments with two model architectures on $29$ target languages from $3$ families show that our suggested approach outperforms all baselines. In particular, it obtains a $31.7\%$ higher absolute accuracy than a previously proposed cross-lingual transfer model and outperforms the previous state of the art by $1.7\%$ absolute accuracy on average over languages.
\end{abstract}

\section{Introduction}
Morphological inflection, an omnipresent phenomenon in many languages, denotes the variation in 
a word's surface form that expresses semantic or syntactic properties like \textit{tense} or \textit{grammatical gender}. 
It yields an abundance of individual word types, while reducing each type's frequency in text. 
The resulting information sparsity is challenging for natural language processing (NLP) systems 
and has lead to the development of approaches for explicit handling of morphology.

\begin{figure}[t]
  \centering
  \includegraphics[width=.90\columnwidth]{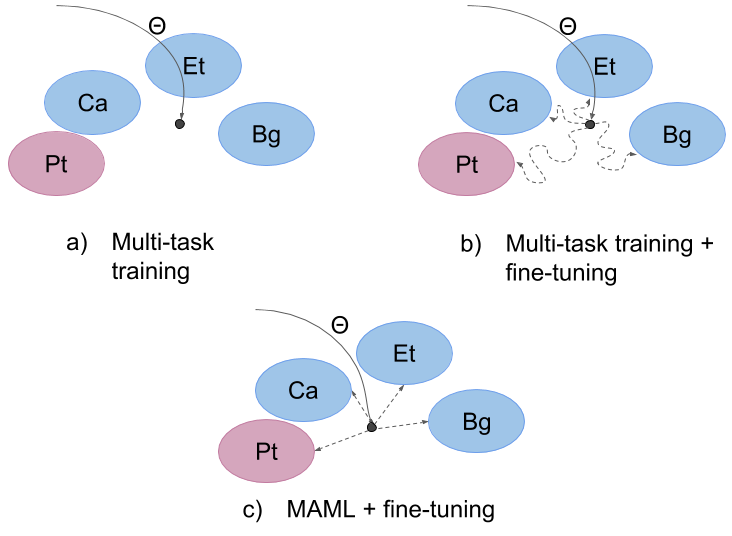}
  \caption{Multi-task training (upper left), multi-task training and fine-tuning (upper right), and MAML and fine-tuning (down). Solid lines represent training of initial parameters, dashed lines fine-tuning. Source languages in blue; target language in purple.}
  \label{fig:MAML}
\end{figure}

The task of morphological inflection consists of automatically generating an indicated inflected form of a given lemma. Over the last years, it has gained popularity in the NLP community and has been featured in a  series of shared tasks \citep{W16-2002,cotterell-EtAl:2017:K17-20,cotterell-EtAl:2018:K18-30}.
Neural sequence-to-sequence (seq2seq) models obtain high accuracy, at least when training sets with many thousands of examples are available
\citep{cotterell-EtAl:2018:K18-30}. 
However, neural models perform poorly if training data is sparse, as is the case for many morphologically rich languages. 
One way to mitigate this problem is via cross-lingual transfer, i.e., by leveraging knowledge from a related resource-rich language. For morphological inflection, knowledge transfer has been shown to be possible via multi-task training on data from several languages, with each language being considered a separate task \citep{kann-cotterell-schutze:2017:Long}. 

In this work, we consider a setting in which we have large training sets from multiple data-rich source languages available, but only a small number of training examples in each of our low-resource target languages. We introduce a novel approach for cross-lingual transfer for morphological inflection, which differs from previous work in two crucial ways: (i) we add a fine-tuning phase to the training process, which enables the model to focus specifically on the low-resource language's examples at the end of training, and (ii) instead of multi-task training, we employ model-agnostic meta-learning (MAML; \citealp{finn2017model}).
Using MAML, we
learn an initial set of model parameters that, instead of just showing good performance on all high-resource tasks, 
is easy to fine-tune on data from new tasks 
with only a few instances. Treating individual languages as separate tasks, our inflection model meta-learns on a set of training languages how to adjust quickly to any new language. Figure \ref{fig:MAML} compares the different approaches.
As a bonus, 
our approach does not require 
training a new model from scratch for each target language:
After a one-time meta-learning phase, language-specific models can be obtained within minutes.

We perform an extensive evaluation of the proposed meta-learned models on data from the 2018 CoNLL--SIGMORPHON shared task on morphological inflection \citep{cotterell-EtAl:2018:K18-30}. In our experiments, we make use of large training sets from $6$ languages belonging to the Romance, Slavic and Uralic language families---Catalan, French, Italian, Bulgarian, Czech, and Estonian---and emulate resource-poor settings for 29 test languages from the same families. Our models outperform the most recent cross-lingual transfer approach, introduced by \citet{kann-cotterell-schutze:2017:Long}, by $31.7\%$ absolute accuracy on average, as well as the state-of-the-art model \citep{makarov-clematide:2018:EMNLP} on the dataset  by $1.7\%$ absolute accuracy on average.  

\paragraph{Contributions.} To summarize, we make the following contributions:
\begin{itemize}
\item We propose a novel approach for cross-lingual transfer for morphological inflection, which is based on MAML.
\item We evaluate our approach on a large set of languages from 3 families and show that it outperforms the previous state of the art.
\item We perform an analysis of the effect of the source languages and find that adding unrelated languages does not hurt the final model, but that languages related to the resource-poor target language are required to obtain improvements.
\end{itemize}

\section{Morphological Inflection}
Many of the world's languages exhibit rich inflectional morphology: The surface form of an individual lexical entry changes in order to express properties such as person, grammatical gender or case. The citation form of a lexical entry is referred to as the \textit{lemma} and the set of all possible surface forms or \textit{inflections} of a lemma as its \textit{paradigm}. 
Each inflection from a paradigm can be associated with a morphological tag, i.e., \texttt{3rdSgPres} is the morphological tag associated with the inflection \textit{writes} of the English lemma \textit{write}.
We display the paradigms of \textit{write} and \textit{eat} in Table \ref{tab:paradigms}.

The presence of rich inflectional morphology is problematic for NLP systems as it greatly increases the token-type ratio in text and, thereby, word form sparsity:
while English verbs can have up to $5$ inflected forms, 
Archi verbs
have thousands \citep{kibrik2017archi}, even by a conservative count.
Thus, an important task in the area of morphology is morphological inflection \citep{durrett-denero:2013:NAACL-HLT,cotterell-EtAl:2018:K18-30}, which consists of mapping a lemma to an indicated inflected form.
An (irregular) English example would be
\begin{align*}
(\textrm{write},  \textrm{\texttt{PAST}}) \rightarrow \textrm{wrote}
\end{align*}
with 
\texttt{PAST} being the target tag, denoting the past tense form.

\begin{table}[t] 
  \setlength{\tabcolsep}{8.5pt}
  \small
  \normalsize
  \centering
  \begin{tabular}{ l @{\hspace*{.8cm}} l l }
    \toprule
    & \textbf{walk} & \textbf{eat} \\
    \midrule
    \texttt{Inf} & { walk} & { eat} \\
    \texttt{3rdSgPres} & { walks} & { eats} \\
    \texttt{PresPart} & { walking} & { eating} \\
    \texttt{Past} & { walked} & { ate} \\
    \texttt{PastPart} & { walked} & { eaten} \\
    \bottomrule
  \end{tabular}
  \caption{\label{tab:paradigms} Paradigms of the English lemmas \textit{walk} and \textit{eat}. While the paradigm of 
  \textit{walk} has only $4$ distinct inflected forms, \textit{eat} has $5$.
  } 
\end{table}
Recently, morphological inflection has frequently been modeled as a sequence-to-sequence problem, where the sequence of target tags and the sequence of input characters constitute the input sequence, and the characters of the inflected word form the output. Neural models define the current state of the art for the task and obtain high accuracy in the high-resource setting. 
In this work, we focus on inflection for resource-poor languages, where state-of-the-art methods are still far from perfect \citep{cotterell-EtAl:2018:K18-30}. 

\paragraph{Formal definition. } 
Let ${\cal M}$ be 
the paradigm slots which are being expressed in a language, and $w$ a lemma in the same language. 
We then define the set of all inflected forms (i.e., the \textit{paradigm}) $\pi$ of $w$ as:
\begin{equation}
  \pi(w) = \Big\{ \big( f_k[w], t_{k} \big) \Big\}_{k \in {\cal M}(w)}
\end{equation}
$f_k[w]$ denotes an inflected form corresponding to tag $t_{k}$, and $w$ and $f_k[w]$ are strings formed by letters from an alphabet $\Sigma$.
The task of morphological inflection consists of predicting a missing form $f_i[w]$ from a paradigm, given the lemma $w$ together with the tag $t_i$.

\section{Meta-Learning}

Meta-learning, or learning-to-learn, 
generally refers to a scenario in which a model learns at two levels.
The first level is task-specific learning, which happens within each training set, and is relatively fast. This learning is further guided by knowledge acquired across tasks via meta-learning, which captures the way in which task structure varies across target domains.

There exist many approaches to meta-learning, e.g., learning a meta-policy for updating model parameters \citep{schmidhuber1987evolutionary}. Within the last years, meta-learning has proven a very successful approach for few-shot 
learning (cf., e.g., \cite{ravi2016optimization}). 
Here, we leverage the model-agnostic meta-learning (MAML; \citealp{finn2017model}) algorithm: we aim at learning a parameter initialization which lends itself easily to adaptation to a new task. 

\subsection{MAML}

The goal of MAML is to train a model 
during a meta-learning 
phase such that it can be easily adapted to a previously unseen task with only a small number of instances. During meta-learning, entire tasks are treated as training examples.

Using MAML, we train a model with parameters $\theta$ on a set of tasks $\{\mathcal{T}_1, \cdots, \mathcal{T}_T\}$ with their associated loss functions $\{\mathcal{L}_1, \cdots, \mathcal{L}_T\}$ and datasets $\{\mathcal{D}_1, \cdots, \mathcal{D}_T\}$. 
For each meta-learning episode, a task $\mathcal{T}_i$ is sampled from a distribution over the available tasks $p(\mathcal{T})$. Then, model training is simulated with $k$ gradient descent steps on $\mathcal{L}_i$, using $K$ examples randomly drawn from $\mathcal{D}_i$: 
\begin{equation}
    \theta' \leftarrow \theta - \eta \nabla_{\theta} \mathcal{L}_{i}(\theta)
\end{equation}
with a learning rate $\eta$.
Subsequently, the model with parameters $\theta'$ is tested on new examples from $\mathcal{D}_i$. The original parameters are then updated by considering how the test error would change with respect to the gradient descent steps applied to $\theta$, using a new learning rate $\eta'$: 
\begin{align}
    \theta & \leftarrow \theta - \eta' \nabla_{\theta} \mathcal{L}_i(\theta') \\
    & = \theta - \eta' \nabla_{\theta} \mathcal{L}_i(\theta - \eta \nabla_{\theta} \mathcal{L}_i(\theta)). \label{eq:4}
\end{align}
Thus, the test error on each $\mathcal{T}_i$ serves as the training error of the meta-learning process. After the meta-training phase, 
task-specific learning is done on a small amount of examples from a new target task, in order to obtain a task-specific model.

\paragraph{Approximating the meta-gradient.} 

Computing the second-order derivative of the loss in Equation~(\ref{eq:4}) is costly.
Since it has been shown that a first-order approximation obtains similar results while achieving a significantly higher computing speed \citep{finn2017model}, we employ the same approximation as \citet{gu2018meta} in practice:
\begin{equation}
    \nabla_{\theta} \mathcal{L}_i(\theta') \approx \nabla_{\theta'} \mathcal{L}_i(\theta')
\end{equation}
The final algorithm we implement for our experiments is shown in Algorithm~\ref{MAML_approx}.

\begin{algorithm}[t]
 \KwIn{Distribution over tasks $p(\mathcal{T})$; \\~ ~ ~ ~ ~ ~ Step hyperparameters $\eta, \eta'$}
 \textit{Random initialization of $\theta$}\;
 \For{$i\leftarrow 1$ \KwTo $n$}{
  Sample tasks $\mathcal{T}_i \sim p(\mathcal{T})$\;
  \For{all $\mathcal{T}_i$}{
      ${\theta_0} \leftarrow \theta$\;
      \For{$j\leftarrow 0$ \KwTo $k-1$}{
          Evaluate $\nabla_{\theta_j} \mathcal{L}(\theta_j)$ with K ~ examples\;
          Compute adapted parameters with ~ GD: ${\theta_{j+1}} \leftarrow \theta_j - \eta \nabla_{\theta_j} \mathcal{L}(\theta_j)$
      }
      Update: ${\theta} \leftarrow \theta - \eta' \nabla_{\theta_k} \mathcal{L}(\theta_k)$
  }
 }
 \caption{First-order approximation of the MAML algorithm.}
 \label{MAML_approx}
\end{algorithm}

\subsection{MAML for Morphological Inflection}

Previous work on cross-lingual transfer has shown that neural models are able to leverage a related language's data for inflection in the limited-data setting \citep{kann-cotterell-schutze:2017:Long}. This suggests that training with MAML could result in even better models, since the algorithm explicitly selects parameters for quick adaptability. 

We consider the following setting: Our goal is to train an inflection model for a resource-poor target language, for which we
only have a limited number of training examples available. Additionally, we have an abundance of data from a set of $n$ high-resource source languages at our disposition. In our main experiments, we consider $n = 6$ and each large dataset consists of 10,000 examples. We further assume that the target language is at least loosely related to at least one source language. We account for this by selecting source and target languages from the same families.\footnote{This is an approximation: Not all languages from the same family are equally similar, and languages from different families can have substantial similarities.
}

We then treat languages as separate tasks and introduce language embeddings which we prepend to each input, following \citet{TACL1081} and \citet{kann-cotterell-schutze:2017:Long}. Task $\mathcal{T}_i$ corresponds to morphological inflection in language $i$.
Meta-training is done on resource-rich languages, and the resulting model is subsequently fine-tuned on the resource-poor language.

\section{Neural Network Models}

As its name suggests, MAML 
can be combined with any model trained with gradient descent. We experiment with two different architectures: a standard attention-based seq2seq model \citep{bahdanau2015neural}, which has been featured in earlier work on multilingual inflection \citep{kann-cotterell-schutze:2017:Long}, as well as a pointer-generator network \citep{gu-EtAl:2016:P16-1}, which has shown to perform well in the low-resource setting \citep{cotterell-EtAl:2018:K18-30}.

\subsection{MED: A Seq2seq Model for Inflection}

Following \citet{kann-cotterell-schutze:2017:Long}, the first model we employ is MED, which stands for \textit{morphological encoder-decoder}. Its architecture is a long short-term memory (LSTM; \citealp{hochreiter1997long}) seq2seq model with attention. We will briefly describe the model here; a more formal overview can be found in \citet{bahdanau2015neural}.

The encoder---the first component of the model---is a bidirectional LSTM: One LSTM encodes the embeddings representing the input sequence
from left to right, and a second LSTM does the same from right to left, yielding the hidden states $\overrightarrow h_i$ and $\overleftarrow h_i$. The final encoder hidden states, which are passed on to the decoder, are  concatenations of the form $h_i = [\overrightarrow h_i, \overleftarrow h_i]$.

The decoder---the second component of our model---is a unidirectional LSTM which is equipped with an attention mechanism: At each time step, a weighted average of the encoder hidden states $h_i$ is computed, depending on the current decoder hidden state $s_t$. This results in a time step-dependent context vector $c_t$. Together with the last predicted output character, $c_t$ is the input to the decoder, 
which predicts the next character. 

The input to the model consists of a concatenation of (i) a tag representing the language, (ii) the morphological tag of the form to be generated, and (iii) the characters of the input lemma. The output is the sequence of characters forming the inflected word.
An example with spaces separating individual elements is: ~\\

\textbf{Input:} ~ \texttt{EN} \texttt{3rd} \texttt{Sg} \texttt{Pres} w r i t e

\textbf{Output:} ~ w r i t e s
~\\

\noindent All elements of the input and output sequences are represented by embeddings.
The language embedding is crucial, since it enables multi-task training on a set of languages.

\subsection{PG: A Pointer-Generator Network}

Second, we experiment with a pointer-generator network architecture \citep{gu-EtAl:2016:P16-1,P17-1099}, which has recently been introduced for morphological inflection by  \citet{sharma-katrapati-sharma:2018:K18-30}. Again, we refer the reader to the original paper for a more detailed explanation and give only a short overview here.

This architecture is also a seq2seq model, but differs from MED in that, instead of always generating characters from a vocabulary, 
it may choose to copy elements from the input over to the output.
The probability of a character $\hat{y}$ is
computed as a sum of the probability of $\hat{y}$ given by the decoder and the probability of copying $\hat{y}$, weighted by the probabilities of generating and copying:
\begin{equation}
    p(\hat{y}) = \alpha p_{\textrm{dec}}(\hat{y}) + (1-\alpha) p_{\textrm{copy}}(\hat{y})
\end{equation}
$p_{\textrm{dec}}(\hat{y})$ is calculated in the same way as for MED, and $p_{\textrm{copy}}(\hat{y})$ depends on the attention weights.
The model computes the probability $\alpha$ with which it generates a new output character as
\begin{equation}
\alpha = \sigma(w_c c_t + w_s s_t + w_y y_{t-1} + b)
\end{equation}
for a context vector $c_t$, a decoder state $s_t$, the embedding of the last output $y_{t-1}$, weights $w_c$, $w_s$, $w_y$, and a bias vector $b$.
The copy mechanism of PG is beneficial in the low-resource setting.

An additional difference from MED is that, instead of encoding tags and characters as one sequence and employing a single encoder, PG uses two encoders: One processes (i) the language tag and (ii) the morphological tags of the target form. The second encodes the characters of the input word. Two attention mechanisms are used, and the concatenation of both context vectors results in the final context vector $c_t$ which is the input to the decoder at time step $t$.

\begin{table*}[t] 
  \setlength{\tabcolsep}{3.4pt}
  \small
  \normalsize
  \centering
  \begin{tabular}{ l || r | r | r | r || r | r | r || r }
    \toprule
    & \multicolumn{4}{|c||}{\textbf{PG}} & \multicolumn{3}{|c||}{\textbf{MED}} &\\
   &  \textbf{MAML-PG} & \textbf{MulPG+FT} & \textbf{MulPG} & \textbf{PG} & \textbf{MAML-MED} 
  & \textbf{MulMED+FT} & \textbf{MulMED} & \textbf{UZH} \\
  \midrule
  \midrule
Asturian & 72.08 & \textbf{73.28} & 66.98 & 69.82 & 68.94 & 65.54 & 58.62 & 71.56 \\
French & \textbf{64.16} & 61.54 & 49.10 & 53.98 & 46.18 & 43.64 & 24.68 & 64.04 \\
Friulian & \textbf{80.00} & 77.00 & 66.00 & 69.40 & 63.40 & 59.20 & 35.40 & 78.20 \\
Italian & 53.12 & \textbf{54.30} & 40.12 & 45.06 & 40.88 & 35.22 & 20.22 & 53.12 \\
Ladin & 58.80 & 58.60 & 40.80 & 66.20 & 50.20 & 47.80 & 31.80 & \textbf{68.60} \\
Latin & 14.04 & 13.90 & 7.90 & 13.36 & 9.10 & 8.28 & 2.52 & \textbf{15.98} \\
Middle French & 83.18 & 80.68 & 65.76 & 81.70 & 71.18 & 66.24 & 40.36 & \textbf{85.16} \\
Neapolitan & 81.00 & 82.20 & 72.60 & 77.00 & 78.00 & 75.20 & 47.40 & \textbf{84.20} \\
Norman & 53.20 & 54.80 & 34.40 & \textbf{58.40} & \textbf{58.40} & 50.00 & 13.20 & 50.40 \\
Occitan & \textbf{77.20} & 72.40 & 66.00 & 70.60 & 70.20 & 70.80 & 46.40 & 74.80 \\
Old French & \textbf{44.48} & 42.86 & 27.82 & 36.16 & 34.58 & 31.78 & 16.84 & 42.14 \\
Romanian & 40.94 & 39.40 & 31.44 & 33.70 & 29.32 & 26.30 & 5.12 & \textbf{42.28} \\
Spanish & 71.62 & \textbf{72.20} & 66.84 & 55.58 & 65.42 & 59.08 & 52.02 & 65.08 \\
Venetian & 74.72 & 73.46 & 68.06 & 75.14 & 71.08 & 67.86 & 40.32 & \textbf{76.40} \\
\midrule
Belarusian & 26.32 & 24.74 & 17.24 & 22.12 & 14.68 & 11.84 & 1.12 & \textbf{27.56} \\
Czech & \textbf{50.36} & 49.08 & 42.86 & 35.46 & 34.46 & 33.06 & 26.14 & 41.56 \\
Kashubian & 59.20 & 58.40 & 56.00 & 57.20 & 58.00 & 59.20 & 19.20 & \textbf{63.60} \\
Lower Sorbian & \textbf{53.52} & 51.54 & 47.22 & 40.72 & 40.36 & 36.88 & 15.36 & 41.88 \\
Old Church Slavonic & \textbf{50.00} & \textbf{50.00} & 35.80 & 47.00 & 38.00 & 29.40 & 7.40 & 42.40 \\
Polish & \textbf{43.40} & 42.60 & 35.38 & 29.62 & 33.40 & 31.56 & 14.36 & 41.82 \\
Russian & \textbf{52.48} & 51.20 & 41.74 & 40.66 & 29.34 & 23.80 & 11.68 & 48.22 \\
Serbo-Croatian & \textbf{39.88} & 37.04 & 23.02 & 35.88 & 26.50 & 21.62 & 9.32 & 38.38 \\
Slovene & \textbf{59.68} & 58.44 & 45.94 & 48.42 & 43.62 & 42.74 & 27.92 & 53.18 \\
Ukrainian & 48.30 & \textbf{48.72} & 43.68 & 36.38 & 35.66 & 28.12 & 18.04 & 46.78 \\
\midrule
Hungarian & 27.64 & 23.22 & 12.46 & 37.78 & 20.02 & 17.60 & 4.04 & \textbf{38.04} \\
Ingrian & \textbf{50.40} & 46.40 & 31.60 & 44.00 & 34.40 & 40.00 & 16.40 & 32.80 \\
Karelian & 85.20 & 84.80 & 62.40 & \textbf{90.80} & 79.60 & 71.20 & 29.20 & 79.20 \\
Livonian & 28.00 & 27.00 & 22.20 & \textbf{32.80} & 22.80 & 22.60 & 2.60 & 29.80 \\
Votic & \textbf{26.60} & 24.60 & 11.20 & 22.60 & 25.60 & 25.20 & 11.40 & 23.00 \\
\midrule
\midrule
\textbf{average} &\textbf{54.12} & 52.91 & 42.50 & 49.23 & 44.60 & 41.44 & 22.38 & 52.42 \\
    \bottomrule
  \end{tabular}
  \caption{\label{tab:results} Inflection  accuracy on the test languages; best results in bold. Languages listed by language family; from top to bottom: Romance, Slavic, Uralic. 
  } 
\end{table*}

\section{Experimental Setup}
\subsection{Data}

To simplify comparison with other work, we experiment on 
a collection of datasets provided by \citet{cotterell-EtAl:2018:K18-30} for the 2018 edition of the CoNLL--SIGMORPHON shared task on morphological inflection. 
For all considered resource-poor settings, we use their \textit{low} datasets, which contain $100$ examples each. For resource-rich languages, we take their \textit{high} datasets, which contain 10,000 examples each.

We limit our experiments to languages which belong to either the Romance, Slavic, or Uralic language family.\footnote{While the Romance and Slavic language families are both Indo-European, we still assume languages within each subfamily to be more closely related.} Two training languages from each are randomly selected among those with a \textit{high} dataset:
Catalan and Galician for Romance, Bulgarian and Slovak for Slavic, and Estonian and Northern Sami for Uralic. Portuguese (Romance), Macedonian (Slavic) and Finnish (Uralic) are our development languages, which we use for hyperparameter tuning, and all other languages in the dataset which belong to either the Romance, Slavic, or Uralic family are used for testing.

\subsection{Baselines}

\paragraph{UZH.} 

Our first baseline is 
the winning system of the low-resource setting of the CoNLL--SIGMORPHON 2018 shared task \citep{makarov-clematide:2018:K18-30}.
It learns a latent alignment between input and output forms which is used for transducing an input to its inflected form and represents the state of the art on our dataset, when averaging over evaluation languages.\footnote{We trained this model with MAML as well, but found it to suffer from the increased character vocabulary which was necessary to account for all languages' alphabets.
}
Since ensembling is orthogonal to our contribution, we compare to the single-model version of UZH.
We use the code and hyperparmeters from \citet{makarov-clematide:2018:EMNLP}.\footnote{\url{https://github.com/ZurichNLP/emnlp2018-imitation-learning-for-neural-} \url{morphology}}

\paragraph{MulMED+FT and MulPG+FT.} 

We further compare to versions of MED and PG which differ from our proposed MAML-based approaches in that they are trained in a multi-task fashion on a set of languages as proposed by \citet{kann-cotterell-schutze:2017:Long}, i.e., there is no meta-learning. For multi-task training, we use the same languages as for MAML, and prepend language embeddings to the input. In contrast to \citet{kann-cotterell-schutze:2017:Long}, we do \textit{not} include the resource-poor language's data during training, but fine-tune the models on those examples. 
These approaches have not been proposed before in the morphological inflection literature and show the effect of meta-learning most directly.

\paragraph{MulMED and MulPG.} 

These baselines are similar to MulMED+FT and MulPG+FT. However, we now present the resource-poor language's data to the models during training and do not have a fine-tuning stage. As a result, these are standard multi-task models. This approach has first been introduced by \citet{kann-cotterell-schutze:2017:Long} for MED, and assumes a slightly different setting as compared to ours, since it requires the target language to be known in advance. For each new language, a model has to be trained from scratch.

\paragraph{PG.} 
We further compare to a plain version of the PG model, without fine-tuning or multi-task training.

\paragraph{MED.} We also experimented with a monolingual MED model, but accuracy was close to zero for all languages. This was an expected outcome, given earlier results \citep{senuma-aizawa:2017:K17-20}. We do, thus, not discuss this baseline in the remaining parts of the paper.

\subsection{Metric} 

We evaluate our inflection models using accuracy at the word level: An inflected form is counted as correct only if it exactly matches the reference.

\subsection{Hyperparameters and Training Regime}
\label{subsec:HP}

\paragraph{MED.} 

For all MED models, we use the hyperparameters suggested by \citet{kann-schutze:2016:P16-2}:\footnote{
\url{http://cistern.cis.lmu.de/med}
}
In particular, the encoder and decoder hidden states are 100-dimensional, and embeddings are 300-dimensional.
For training, we use Adadelta \citep{zeiler2012adadelta} with a batch size of $20$.

\paragraph{PG.} 

We use the hyperparameters suggested by \citet{sharma-katrapati-sharma:2018:K18-30}:\footnote{\url{https://github.com/abhishek0318/ conll-sigmorphon-2018}} Both our hidden states and embeddings are $100$-dimensional. We use Adam \citep{kingma2014adam} for training and dropout \citep{srivastava2014dropout} with a probability parameter of $0.5$.

\paragraph{MAML and training.} 

We tune MAML-related hyperparameters on the development languages. Namely, we search for an appropriate amount of training epochs and a good $k$ value, i.e., the number of update steps to make per set of examples during simulated training. We experiment with $k \in \{2, 5, 8\}$
and obtain an accuracy of $61.99$, $62.27$, and $61.62$, respectively, when averaged over all languages and $5$ models per language. Thus, we choose $k=5$ for our final experiments.

Multi-task training or training with MAML is carried out for $60$ epochs for all model architectures. We fine-tune for at least 300 epochs and, for PG, extend by $100$ epochs each time the results on the development set have improved within the last $100$ epochs \citep{sharma-katrapati-sharma:2018:K18-30}. 
Monolingual MED and PG models are trained for 300 epochs. 
In all cases, we evaluate accuracy on the development set after every epoch and use the best model for testing.

\begin{table*}[t] 
  \setlength{\tabcolsep}{6.5pt}
  \small
  \normalsize
  \centering
  \begin{tabular}{ l || c | c | c | c || c | c | c | c }
    \toprule
    & \multicolumn{4}{|c||}{\textbf{MAML-PG}} & \multicolumn{4}{|c}{\textbf{MulPG+FT}} \\
    & \textbf{ALL} & \textbf{LF} & \textbf{OtherLF} & \textbf{SINGLE} & \textbf{ALL} & \textbf{LF} & \textbf{OtherLF} & \textbf{SINGLE} \\
    \midrule
    Portuguese & 84.62 & \textbf{86.96} & 61.64 & 60.10 & 85.00 & \textbf{87.56} & 58.00 & 60.10 \\
    Macedonian & 70.28 & \textbf{70.62} & 53.94 & 56.86 & 69.56 & \textbf{70.98} & 55.92 & 56.86 \\
    Finnish & 27.58 & \textbf{27.96} & 14.82 & 18.58 & \textbf{26.74} & 26.00 & 13.32 & 18.58 \\
    \midrule
    \textbf{average} & 60.83 & \textbf{61.85} & 43.47 & 45.19 & 60.43 & \textbf{61.51} & 42.41 & 45.19 \\
    \bottomrule
  \end{tabular}
  \caption{\label{tab:results_LR2} Accuracy for LF and OtherLF models, as well as monolingual models (SINGLE) and models trained on all 6 source languages (ALL). Averaged over 5 runs; best result for each model type and language in bold.
  } 
\end{table*}

\section{Results and Discussion}

The results for all languages, grouped by language family, are shown in Table \ref{tab:results}. We make the following observations:
\begin{itemize}
    \item For both MED and PG, MAML-trained models outperform all other models of the same architecture: MAML-PG (resp. MAML-MED) obtains a $1.21\%$ (resp. $3.16\%$) higher accuracy on average over languages than the second best model MulPG+FT (resp. MulMED+FT). This demonstrates that MAML is more effective than multi-task training.
    
    \item For both architectures, models which are obtained by multi-task training and subse1uent fine-tuning outperform models trained exclusively in a multi-task fashion: MulPG+FT (resp. MulMED+FT) obtains a $10.41\%$ (resp. $19.06\%$) higher accuracy on average than MulPG (resp. MulMED). Since differences in performance are substantial, we conclude that the use of fine-tuning---with or without MAML---is important.
    
    \item All PG models outperform their corresponding MED models. This is in line with the findings of \citet{cotterell-EtAl:2018:K18-30} that the pointer-generator network is a strong model for morphological inflection in the low-resource setting.
    
    \item Plain PG performs worse than fine-tuned PG models, but better than MulPG. That both MAML-PG and MulPG+FT outperform PG shows the importance of fine-tuning. Multi-task training as suggested by \citet{kann-cotterell-schutze:2017:Long} seems to not work well in our setup. One possible reason for this is that we include languages from 3 different families as opposed to the original experiments which focussed on one family at a time. Thus, better ways to account for the amount of unrelated examples are probably needed for MulPG.

    \item MulPG+FT and MAML-PG perform better than UZH, the state-of-the-art model. We conclude that cross-lingual transfer is a promising direction to improve morphological inflection for resource-poor languages.
    
    \item Looking at individual languages, MAML-PG performs better than the plain PG model in all cases, except for 3 Romance (Ladin, Norman, and Venetian) and 3 Uralic (Hungarian, Karelian, and Livonian) languages. There are \textit{no} exceptions for the Slavic family.
    That relatively many monolingual models outperform cross-lingual transfer models for Uralic languages
    might be due to these languages being less similar than 
    those in the other families.
    We look into this in more detail in the next section.
\end{itemize}

\section{Impact of the Source Languages}

The goal of this work is to show that we can use MAML to train a single model which can be easily adapted to novel languages from the same language families. The degree to which this is possible depends on the availability of suitable source languages. This section addresses three questions about possible source languages:

\paragraph{Q1.} 
Is using unrelated languages for multi-task training or training with MAML harmful?

\paragraph{Q2.} How much does performance decrease
if no data from a language similar to the target language is used for initial training?

\paragraph{Q3.} Do the answers for the above two questions differ, depending on if we use MAML or multi-task training?

\paragraph{}In order to find answers to these questions, we train the following additional MulPG+FT and MAML-PG models: 
First, we train models on source languages from all but the language family of the target language (OtherLF), e.g., a model for Portuguese  is trained on Slavic and Uralic source languages only.
Second, we train models exclusively on source languages from the target language's family (LF), e.g., a model for Portuguese is trained on Romance source languages only.

We evaluate on our development languages Portuguese (from the Romance language family), Macedonian (from the Slavic language family), and Finnish (from the Uralic language family).

\subsection{Results and Discussion}

The results for the LF and OtherLF models are shown in Table \ref{tab:results_LR2}.
We observe the following:

\paragraph{Q1.} Looking at the difference between ALL and LF, we find that model accuracy does not benefit much from excluding unrelated languages: The average performance increases only slightly compared to the overall gains for both MAML-PG and MulPG+FT. Given that in practice it might be difficult to accurately judge the similarity of two languages with respect to the morphological inflection task, it is valuable to know that training a highly multilingual model only hurts performance slightly.

\paragraph{Q2.} Considering the question of how a model performs if we do not include any closely related high-resource language during training at all, 
we can see from Table \ref{tab:results_LR2} that \textit{excluding} source languages from a target language's family 
strongly decreases model performance for all target languages.
For Macedonian and Finnish, results for OtherLF for both MAML-PG and MulPG+FT are even worse than for SINGLE. 
Thus, it might be better to train a monolingual model than a multilingual model with only 
unrelated 
languages.

\paragraph{Q3.} The answer to Q1 is the same for both MAML-PG and MulPG+FT, i.e., the training algorithm does not influence whether adding additional languages hurts performance. 
Considering Q2 and looking at Table \ref{tab:results_LR2}, we see that performance decreases minimally more if no related language data is available for a given target language when using multi-task training.

\section{Related Work}
\paragraph{Morphological inflection. } 
Recently, neural seq2seq models have gained popularity for  morphological inflection \citep{faruqui-EtAl:2016:N16-1,kann-schutze:2016:P16-2,aharoni-goldberg:2017:Long,bergmanis-EtAl:2017:K17-20}.
Our work extends the line of research on morphological inflection for \textit{resource-poor languages}. The CoNLL--SIGMORPHON 2017 and 2018 shared tasks \citep{cotterell-EtAl:2017:K17-20,cotterell-EtAl:2018:K18-30} both featured settings with small amounts of training data, encouraging the development of systems for such cases, including \citet{sharma-katrapati-sharma:2018:K18-30} and \citet{makarov-clematide:2018:K18-30}. 
However, we are the first to apply a meta-learning approach to the task of morphological inflection. 
Most similar to our work is \citet{kann-cotterell-schutze:2017:Long}, which introduces multilingual inflection models. However, as just mentioned, the authors do not use meta-learning, and further assume that the training data in the resource-poor language is available for the entire training of the model. Here, we relax this assumption and train a general model first, which can then be quickly and easily adapted. However, we compare to multilingual models---both with the architecture used in that previous work as well as for a more recent one---as a baseline.

Non-neural approaches for morphological inflection include \citet{durrett-denero:2013:NAACL-HLT}, \citet{eskander-habash-rambow:2013:EMNLP}, and \citet{nicolai-cherry-kondrak:2015:NAACL-HLT}.

\paragraph{Meta-learning. } 
Meta-learning \citep{bengio1990learning,naik1992meta}---or learning to learn---dates back decades, but has recently become a more active research topic.
Meta-learning techniques have been applied to a variety of machine learning problems. Outside of NLP, successful applications include image recognition \citep{ravi2016optimization,vinyals2016matching}
and video object segmentation \citep{yang2018efficient,behl2018meta}.
Within NLP, meta-learning approaches are more scarce. However, MAML has been employed for training machine translation \citep{gu2018meta} or semantic parsing \citep{huang2018natural}, \textit{inter alia}. 
To the best of our knowledge, we are the first to use a meta-learning approach for the training of neural morphological generation models.

\paragraph{Cross-lingual transfer learning. } Our work is further
related to work on cross-lingual transfer learning \citep{wu1997stochastic,yarowsky2001inducing}, which has been explored for many NLP tasks, including parsing \citep{sogaard:2011:ACL-HLT20112,naseem-barzilay-globerson:2012:ACL2012}, speech recognition \citep{huang2013cross}, machine translation \citep{ha2016toward,TACL1081}, or, as mentioned previously, morphological inflection \citep{kann-cotterell-schutze:2017:Long}. Our work is similar to \citet{TACL1081} and \citet{kann-cotterell-schutze:2017:Long} in that we augment a neural seq2seq model's input with an explicit encoding of the language. 
Here, we show that it is beneficial to use meta-learning to find a good parameter initialization for fine-tuning.

\section{Conclusion and Future Work}
We propose to use MAML to train an inflection model on a set of data-rich source languages such that it is easily adaptable to a new target language. Language-specific models can then be obtained via fine-tuning on small amounts of new data.
Experiments with an attention-based seq2seq architecture and a pointer-generator network on $29$ target languages from $3$ different families showed that our suggested method outperforms 
all baselines.
Furthermore, our MAML-trained models obtained a $1.7\%$ higher absolute accuracy on average over all languages than the previous state of the art.
In future work, we hope to investigate how MAML can be applied to other morphological tasks such as morphological segmentation and look into using inflection models to improve systems for downstream applications for resource-poor languages.

\section{Acknowledgments}
This work has received support from Samsung Advanced Institute of Technology (\textit{Next Generation Deep Learning: from Pattern Recognition to AI}) and Samsung Electronics (\textit{Improving Deep Learning using Latent Structure}). It further benefited from the donation of a Titan V GPU by NVIDIA Corporation.

\fontsize{9.8pt}{10.8pt} \selectfont
\bibliography{AAAI}
\bibliographystyle{aaai}
\end{document}